\documentclass[letterpaper, 10 pt, conference]{ieeeconf} 
\IEEEoverridecommandlockouts

\usepackage{amsmath}
\usepackage{graphicx}
\usepackage{amssymb}
\usepackage{multirow}
\usepackage{tikz}
\usepackage{adjustbox}
\usepackage{booktabs}

\title{\LARGE \bf Temporal Convolutions for Multi-Step Quadrotor Motion Prediction}

\author{
  Samuel Looper$^{1}$ and Steven L. Waslander$^{2}$
  \thanks{$^{1}$Engineering Science, University of Toronto, Canada, \texttt{samuel.looper@mail.utoronto.ca}} \\
  \thanks{$^{1}$Institute for Aerospace Studies, University of Toronto, Canada, \texttt{stevenw@utias.utoronto.ca}} \\
}

\begin{document}
\maketitle


\begin{abstract}
Model-based control methods for robotic systems such as quadrotors, autonomous driving vehicles and flexible manipulators require motion models that generate accurate predictions of complex nonlinear system dynamics over long periods of time. Temporal Convolutional Networks (TCNs) can be adapted to this challenge by formulating multi-step prediction as a sequence-to-sequence modeling problem. We present End2End-TCN: a fully convolutional architecture that integrates future control inputs to compute multi-step motion predictions in one forward pass. We demonstrate the approach with a  thorough analysis of TCN performance for the quadrotor modeling task, which includes an investigation of scaling effects and ablation studies. Ultimately, End2End-TCN provides 55\% error reduction over the state of the art in multi-step prediction on an aggressive indoor quadrotor flight dataset. The model yields accurate predictions across 90 timestep horizons over a 900 ms interval.  

\end{abstract}



\section{Introduction}

While  autonomous robotic systems offer tremendous potential benefits in a wide range of commercial operations, their safe operation will require highly accurate localization and control methods for collision avoidance and action execution. Model-based state estimation and control have demonstrated strong performance and robustness across the operational domain of remote aircraft~\cite{greeff2018flatness,alexis2012model,bouffard2012learning}, autonomous vehicles~\cite{kong2015kinematic,grigorescu2020survey} and flexible manipulators ~\cite{liu2018dynamic}, to name a few. As such, dynamic system modeling is critical to the effort of developing safe autonomous robotic systems that can perform precise motions throughout their operating envelopes. 

As a primary motivating example, in this work we focus on multi-step prediction for quadrotor UAVs.  Indeed, developing models of quadrotor flight solely from first principles has proven to be a challenge. Quadrotors are underactuated systems whose translational dynamics are tightly coupled with highly nonlinear rotational dynamics. In real-world environments, aerodynamics, motor dynamics, and asymmetrical mass distributions can be significant disturbances, but are often poorly characterized in most physics-based quadrotor models~\cite{hoffmann2011precision}. 

Another line of research focuses on developing statistical quadrotor models from measured flight data. Specifically, discrete time neural network designs have shown the greatest promise in modeling complex quadrotor dynamics due to their strong expressive power. A recent work to benchmark neural network models on quadrotor state prediction performance employ Recurrent Neural Network (RNN) models to sequentially learn time-correlated features in quadrotor state telemetry time series data~\cite{mohajerin2019multistep}. While these models have demonstrated state-of-the-art performance, they have several limitations. The sequential nature of these models leads to longer computation times due to the lack of parallelization, and can cause unstable gradients at training time~\cite{pascanu2013difficulty}. Furthermore, current models are limited in their ability to learn time-correlated features over long time horizons~\cite{vaswani2017attention}.

Temporal convolution-based architecture provide a potential solution to the limits of RNNs for the task of quadrotor state modeling. Temporal Convolutional Networks (TCNs) have demonstrated the ability to accurately model time series in a variety of contexts~\cite{oord2016wavenet},\cite{bai2018empirical},\cite{lea2017temporal} and have the potential to provide a sparse an efficient model able to learn features over long time histories. In this work, we apply TCNs to a discrete time multi-step series forecasting problem, which we adapt to the non-autonomous dynamics of robotic systems. This allows for TCN models to be trained and evaluated on indoor quadrotor flight telemetry. 

Thus, in this paper, we perform the first in-depth study of convolution-based architectures for quadrotor modeling. We present End2End-TCN: a novel method of applying TCNs to robotic system modeling by integrating the control input into the input state vector.  This model surpasses the current state of the art and several alternative models in prediction accuracy, generating useful future state predictions over longer periods of time for the purpose of model-based control and state estimation. We perform a comprehensive series of experiments to characterize the performance of TCNs with respect to model size and past state history length. We further provide an analysis of prediction samples and error distributions to characterize model performance. Most importantly, we demonstrate that a TCN-based model can provide a memory-efficient representation of quadrotor dynamics and yield a 55\% reduction in prediction error over a 900 ms interval.

\section{Related Works}
\label{sec:literature}
\textbf{Empirical Methods.} As a result of the success of model-based quadrotor control methods, the dynamics of quadrotor flight have been extensively studied in literature. Previous research that developed quadrotor test bed platforms~\cite{hoffmann2004stanford}, developed dynamical system models~\cite{bouabdallah2007full},\cite{pounds2006modelling} and characterized significant aerodynamic effects~\cite{mahony2012multirotor} have laid the foundation for a principled approach to developing quadrotor models. In these works, simplified models of quadrotor geometry, rotor thrust, and aerodynamics were used to derive equations of motion. Such physics-based models have been further refined by deriving more complex aerodynamic models~\cite{hoffmann2011precision} or by using blade element momentum theory~\cite{barcelos2018performance}~\cite{bauersfeld2021neurobem} to better characterize motor thrust. While many such models obtain parameter values through empirical measurement or offline system identification, recent works have used online parameter estimation to refine their physics-based models over time~\cite{bohmfilter}~\cite{burri2018framework}~\cite{wuest2019online}. 

\textbf{Neural Networks.} Neural networks, on the other hand, provide powerful and flexible statistical models that can model highly complex time-varying phenomena. In the field of statistical rotorcraft flight modeling, early work by Punjabi and Abbeel~\cite{punjani2015deep} showed significant success in learning a nonlinear acceleration model for helicopter flight by training a simple artificial neural network on past flight telemetry, while others ~\cite{bansal2016learning} learned a simpler linearizable model for LQR control. Such models may successfully learn a latent representation of flight data, but are not designed to specifically learn time-correlated features, which have been demonstrated to improve performance in sequence domain tasks. On need look no further than the field of stock price modeling, where early artificial neural networks~\cite{wanjawa2014ann} were quickly surpassed by LSTM models~\cite{zhao2017time} and TCNs~\cite{deng2019knowledge} spcifically due to their ability to learn time-correlated features.

\textbf{Sequence Modeling.} In recent years, deeper networks with new neural network architectures have led to major breakthroughs in sequence modeling. Much of this research has focused on Recurrent Neural Networks (RNNs). Mohajerin et al. leveraged recurrent architectures towards quadrotor modeling by training RNNs with Long-Short-Term Memory gated units on an indoor quadrotor dataset, which greatly improved prediction accuracy for future flight trajectories \cite{mohajerin2019multistep}. This sequential approach mirrors the way discrete dynamical system models are integrated forward in time. However, the ability of an RNN to model time-varying phenomena is limited by the size of its hidden state representation \cite{zhang2016architectural}, and RNN performance degrades significantly as time horizons extend \cite{vaswani2017attention}, both of which limit their usability for quadrotor flight modeling. RNNs also have limitations that make them ill-suited for online robotics applications. They are less computationally efficient than convolution-based architectures that can leverage parallel computation hardware \cite{bai2018empirical} due to the cost of processing time series in a sequential method. Furthermore, RNNs can be challenging to train due to backpropagation through time, which can lead to gradient instability \cite{pascanu2013difficulty}. 

\textbf{Temporal Convolutional Networks.} While RNNs were the dominant approach for time series predictive modeling~\cite{salinas2020deepar},\cite{abdel2019accurate},\cite{shi2017deep}, convolution-based approaches have emerged recently as a viable alternative. Early work by van den Oord et al. on WaveNet \cite{oord2016wavenet} introduced the causal convolution, which modified the standard discrete convolution operation to maintain the temporal structure of time series inputs. Dilated convolutions can be employed to make predictions over large, fixed time horizons and the resulting network can be parallelized for computational efficiency. This results in sparse networks that learn time-correlated features in an efficient and deterministic manner, which are called Temporal Convolutional Networks (TCNs). 

Studies have shown that TCNs outperform recurrent networks across a wide range of sequence modeling tasks \cite{bai2018empirical}. TCNs were further explicitly applied to time series modeling by Borovykh et al. \cite{borovykh2017conditional}. More relevant to quadrotor modeling, TCNs were used in action segmentation tasks\cite{lea2017temporal} and were combined with Empirical Mode Decomposition (EMD) to predict dynamic climate effects \cite{yan2020temporal}. These prior works demonstrate that TCNs have the ability to learn temporal patterns in robotic motion over long periods and model highly complex dynamical systems.

Many applications of deep learning in robotics learn temporal patterns by simply concatenating images or system state inputs \cite{wang2017deepvo}. However, this only works over short time periods. Recent work by Kaufmann et al. leveraged TCNs to process sensor input information in an end-to-end learning-based architecture for quadrotor control \cite{kaufmann2020deep}. While this study demonstrates the utility of TCNs in the context of quadrotor state information processing, there is still a clear lack of research on the ability of TCNs to explicitly model robotic systems over a long horizon of future state predictions.


\section{Problem Formulation}
\label{sec:formulation}

By treating quadrotor flight dynamics as a time series predictive modeling problem, we can perform sequence-to-sequence modeling to learn a function that can predict future states. We first define a parameterization of the quadrotor state, $\mathbf{x}(t) = \begin{bmatrix} \mathbf{\eta}(t) & \mathbf{r}(t) & \mathbf{\xi}(t) & \mathbf{v}(t) \end{bmatrix} \in \mathbb{R}^n$, which includes position, $\mathbf{r}(t) \in \mathbb{R}^3$, and velocity, $\mathbf{v}(t)\in \mathbb{R}^3$, in a world frame, $\mathcal{F}_w$, orientation,  $\mathbf{\eta}(t) \in SO(3)$, represented by Euler rotation angles from a body frame, $\mathcal{F}_b$, about axes XYZ to the world frame, $\mathcal{F}_w$, and rotation rate, $\mathbf{\xi}(t) \in \mathbb{R}^3$, represented by the time derivative of XYZ Euler angles with respect to the body frame, $\mathcal{F}_b$. The diagram below denotes the world frame, $\mathcal{F}_w$, and body frame, $\mathcal{F}_b$, with respect to the quadrotor’s geometry. The way the geometry and reference frames are denoted is based on a quadrotor X-configuration, where the roll and pitch axes are offset by 45 degrees from the rotor arms.  

\begin{figure}[thpb]
  \centering
  \includegraphics[width= 0.48\textwidth]{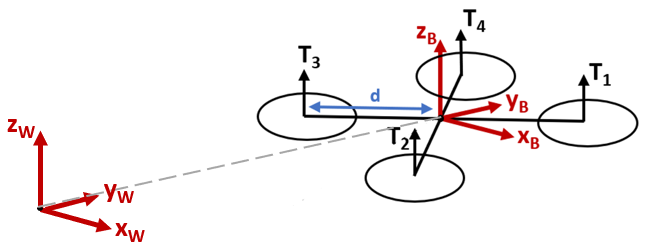}
  
  \caption{Model of a quadrotor as a rigid body from a body and inertial frame.
  }
  \label{fig:quadrotor_model_fig}
\end{figure}

\begin{figure*}[thpb]
  \centering
  \includegraphics[width= 0.8\textwidth]{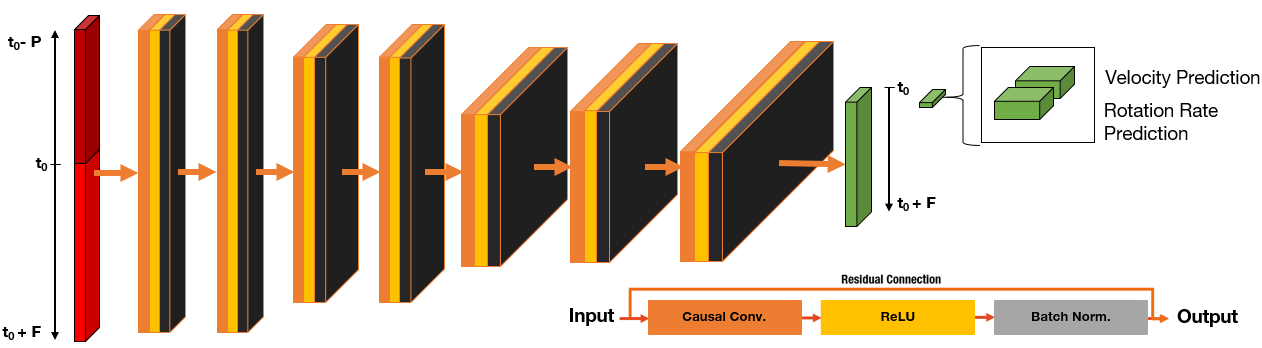}
  
  \caption{Full End2End-TCN architecture.}
  \label{fig:e2e_full}
\end{figure*}
\newpage
This state represents the quadrotor’s pose with 6 degrees of freedom (given the orientation representation) and a measure of its first rate of change. The full system is further characterized by a control input, $\mathbf{u}(t) = \begin{bmatrix} u_1(t) & u_2(t) & u_3(t) & u_4(t) \end{bmatrix} \in \mathbb{R}^d $, representing four motor commands  which are generated by the quadrotor’s controller and linearly map to desired rotor speeds, $\mathbf{\varpi} = \begin{bmatrix} \varpi_1(t) & \varpi_2(t) & \varpi_3(t) & \varpi_4(t) \end{bmatrix} \in \mathbb{R}^d$.

In this discrete quadrotor state formulation, we consider a dynamic system represented by the function $f \colon \mathbb{R}^{n} \to \mathbb{R}^{n}$ that maps a past state representation, $\mathbf{x_{t_0-1}}$, to a future state representation, $\mathbf{x_{t_0}}$, and a function  $g \colon \mathbb{R}^{n} \to \mathbb{R}^{m}$ that maps a state representation, $\mathbf{x_{t_0}}$, to a state observation, $\mathbf{y_{t_0}}$. In the non-autonomous case, the function $f$ maps both the past state, $\mathbf{x_{t_0-1}}$, and a control input, $\mathbf{u_{t_0}} \in \mathbb{R}^d$, to the state observation, $\mathbf{y_{t_0}}$. 

However, to fully leverage the ability of convolutional neural networks to compute state predictions in parallel, we extend this formulation to a multi-step prediction case of length $F$. Note that in the non-autonomous case, past and future control inputs will be required as inputs to this function, as the future state, $\mathbf{x_{t_0+i}}$, is dependent on the future control input, $\mathbf{u_{t_0+i}}$. Furthermore, given the complexity of dynamic effects such as aerodynamics on quadrotor motion, the state parameterization above may not meet the Markov condition. Thus, we theorize that prediction accuracy will be improved by providing a sequence of $P$ input states. As such, we seek to model the function $f^{(P, F)}$ mapping a series of $P$ past states, $P$ past control inputs, and $F$ future control inputs to the series of $F$ future states. Note that this model assumes access to the full state representation, which is only possible in the case of weak nonlinear observability.  

Modeling this discrete function can thus be formulated as a sequence-to-sequence modeling problem. We consider a sequence of prior system states, $ \mathbf{X}_p = \begin{bmatrix} \mathbf{x}_{t_0-P+1} & \mathbf{x}_{t_0-P+2} & ... & \mathbf{x}_{t_0} \end{bmatrix}  \in \mathbb{R}^{n \times P}$, prior control inputs, $ \mathbf{U}_p = \begin{bmatrix} \mathbf{u}_{t_0-P+1} & \mathbf{u}_{t_0-P+2} & ... & \mathbf{u}_{t_0} \end{bmatrix} \in \mathbb{R}^{d \times P}$, and future control inputs, $ \mathbf{U}_f = \begin{bmatrix} \mathbf{u}_{t_0+1} & \mathbf{u}_{t_0+2} & ... & \mathbf{u}_{t_0+P} \end{bmatrix} \in \mathbb{R}^{d \times F}$, and seek to estimate future system states, $ \mathbf{X}_f = \begin{bmatrix} \mathbf{x}_{t_0+1} & \mathbf{x}_{t_0+2} & ... & \mathbf{x}_{t_0+F} \end{bmatrix} \in \mathbb{R}^{n \times F}$. Thus, given a sequence-to-sequence function $\hat{f} \colon \mathbb{R}^{(n+d)\times P} \to \mathbb{R}^{n\times F}$ generating a future system state prediction $ \mathbf{\hat{x}}_f = \begin{bmatrix} \mathbf{\hat{x}}(t_0+1) & \mathbf{\hat{x}}(t_0+2) & ... & \mathbf{\hat{x}}(t_0+F) \end{bmatrix}$, we can use statistical methods to minimize a reconstruction loss $\ell(\mathbf{\hat{x}}, \mathbf{x})$ over a set of known future quadrotor states.
\begin{equation}
    \mathcal{L} = \frac{1}{F} \sum^{i=t_0+F}_{i=t_0} \ell(\mathbf{\hat{x}}_{i}, \mathbf{x}_{i})
\end{equation}

\section{Methodology}
\label{sec:methodology}

Given historical quadrotor state data, neural network model inputs and labels are generated in a semi-supervised manner. As per the problem formulation, model inputs include prior quadrotor states $\mathbf{X}_p$, control inputs $\mathbf{U}_p$, and future control inputs $\mathbf{U}_f$. The sample labels $\mathbf{Y}_f$ correspond to a series of truncated quadrotor states, $\mathbf{y}_k = \begin{bmatrix} \mathbf{\xi}_k & \mathbf{v}_k \end{bmatrix}$, which include the translational and rotational velocities from time $t_0+1$ to time $t_0+F$. 

 A fully convolutional neural network model, dubbed End2End-TCN, is trained on this time series data to provide quadrotor state predictions over $F$ time steps. Crucially, in order to make multiple predictions over this non-autonomous dynamical system, past and future control input must be integrated into the discrete sequence modeling problem formulation. End2End-TCN integrates this information into a fixed sequence length input, $\mathbf{\tilde{X}} = \begin{bmatrix}\mathbf{\tilde{X}}_p & \mathbf{\tilde{X}_f} \end{bmatrix} \in \mathbb{R}^{(n+d) \times (P+F)}$, composed of augmented states, $\mathbf{\tilde{x}_k} = \begin{bmatrix} \mathbf{\eta}_k & \mathbf{r}_k & \mathbf{\xi}_k & \mathbf{v}_k & \mathbf{u}_k\end{bmatrix} \in \mathbb{R}^{(n+d)}$, for prior states ($k < P$) and, $\mathbf{\tilde{x}_k} = \begin{bmatrix} \mathbf{0} & \mathbf{0} & \mathbf{0} & \mathbf{0} & \mathbf{u}_k\end{bmatrix} \in \mathbb{R}^{(n+d)}$, for future states ($k \geq P$).
 
\begin{figure}[thpb]
  \centering
  \includegraphics[width= 0.48 \textwidth]{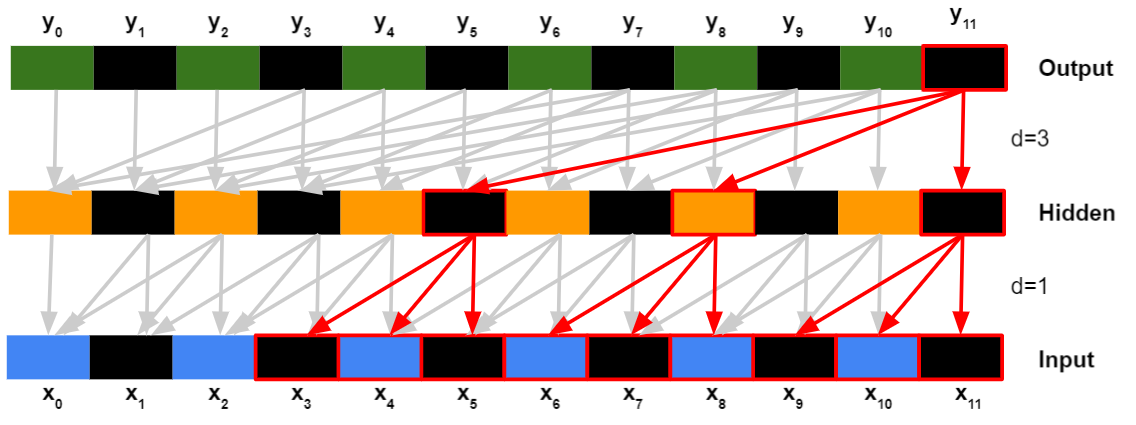}
  
  \caption{Causal convolutions over a series of layers with exponentially increasing dilation factor.}
  \label{fig:tcn_layer}
\end{figure}

The model is built on a series of causal convolutions, as first developed in \cite{oord2016wavenet}, and as implemented in \cite{bai2018empirical}. To achieve the desired effect, a causal convolution block is composed of a series of causal convolutions with dilations that increase exponentially at every layer, as depicted in figure~\ref{fig:tcn_layer}.

Causal convolution blocks are stacked with a nonlinear activation function and batch normalization, with a residual connection applied for gradient stability. These blocks are stacked in a layered architecture to form a deep, overparameterized neural network as in \cite{lea2017temporal} and \cite{chen2020probabilistic} (see figure~\ref{fig:e2e_full}). End2End-TCN was designed to output a full time series of predicted states at every forward pass, allowing for simultaneous multi-step prediction of quadrotor states. 

\subsection{Physics-based Model} \label{physics}
A key part of the study of TCNs for quadrotor modeling is ascertaining whether prior knowledge of the system's dynamics is required to improve prediction accuracy. Consequently, we develop a physics-based model of quadrotor flight derived for the AsTec Pelican flights in the test set. This model is based on a simplified wire-frame model of the quadrotor as per figure \ref{fig:quadrotor_model_fig}, which is represented by four arms with a uniform mass and a length $d$. For the specified platform, the arms form a right angle with one another. Fixed to each arm is a rotor, which is modeled by a point mass generating a longitudinal thrust $T_i$ and rotational torque $Q_i$. The body frame is defined such that the rotors lie on the XY plane, the x-axis points in the direction directly between the first and second rotors, and the z-axis points in the direction of the torques generated by any individual rotor. The diagram in figure \ref{fig:quadrotor_model_fig} depicts the wireframe quadrotor model and the two corresponding reference frames (inertial and body). 
The complex motor and rotor dynamics are approximated by a quadratic relationship between the rotor angular velocity,  $\varpi_i \in \mathbb{R}$, in its discrete representation. This is based on the rotor dynamic equation in stead state with a freestream velocity of zero \cite{mahony2012multirotor}, which can be parameterized with respect to a thrust coefficient $c_T$, the density of air $\rho$, the rotor radius $r_i$ and the rotor area $ A_{r,i}$.   

\begin{equation}
T_i = c_T \, \rho \, A_{r,i} \, r_i^2 \,\varpi_i^2 = C_T \, \varpi_i^2
\end{equation}

\begin{equation}
Q_i = C_Q \, \varpi_i^2
\end{equation}

The total thrusts and torques can thus be calculated from individual rotor contributions in the vectorized equation below.  

\begin{equation}
\begin{bmatrix}
T_{tot}\\ 
\tau_1\\ 
\tau_2\\ 
\tau_3
\end{bmatrix} = C_T \begin{bmatrix}
1 & 1 & 1 & 1\\ 
\frac{d}{\sqrt{2}} & \frac{-d}{\sqrt{2}} & \frac{d}{\sqrt{2}} & \frac{-d}{\sqrt{2}}\\ 
\frac{-d}{\sqrt{2}} & \frac{-d}{\sqrt{2}} & \frac{d}{\sqrt{2}} & \frac{d}{\sqrt{2}}\\ 
\frac{-C_Q}{C_T} & \frac{C_Q}{C_T} & \frac{-C_Q}{C_T} & \frac{C_Q}{C_T}\\ 
\end{bmatrix} 
\begin{bmatrix}
\varpi_1^2\\ 
\varpi_2^2\\ 
\varpi_3^2\\ 
\varpi_4^2
\end{bmatrix}
\end{equation}

For state derivatives, we reference a quadrotor state in the form $\mathbf{x}(t) = \begin{bmatrix} \mathbf{\eta}(t) & \mathbf{r}(t) & \mathbf{\xi}(t) & \mathbf{v}(t) \end{bmatrix} \in \mathbb{R}^n$ as per the problem formulation in section 3. The derivative of position is trivial, $\mathbf{\dot{r}(t)} = \mathbf{v}(t)$. The orientation derivative can be obtained from the body rotation rates with an additional coordinate transform $\mathcal{F}_b \to \mathcal{F}_w$ in matrix form ($\mathbf{R_b^i}$). 
\begin{equation}
\mathbf{\dot{\eta}} = \begin{bmatrix}
\dot{\theta} \\ 
\dot{\phi}\\ 
\dot{\psi}
\end{bmatrix} = \mathbf{R_b^i} \mathbf{\omega} = 
\begin{bmatrix}
1 & 0 & - \sin(\theta) \\ 
0 & \cos(\phi) & \sin(\phi) \cos(\theta)\\ 
0 & -\sin(\phi) & \cos(\phi) \cos(\theta)
\end{bmatrix}  \mathbf{\omega}
\end{equation}

Translational acceleration can be written with respect to the force and torque from equation 4 using Newton’s 2nd Law. Motor thrust $\mathbf{f_T}$ is transformed from $\mathcal{F}_b$ to $\mathcal{F}_w$, and additional inertial accelerations due to gravity ($g$) and translational drag $k_t \mathbf{v}$. Lastly, rotational acceleration can be written from Euler’s Equations of Rotational Motion, with a body frame rotor torque $\mathbf{\tau}$ and rotational drag $k_R \mathbf{\omega}$. 
\begin{equation}
\mathbf{\dot{v}} = \frac{1}{m}\mathbf{R_b^i} \mathbf{f_T} - \mathbf{g} - k_t \mathbf{v}
\end{equation}

\begin{equation}
\mathbf{I} \mathbf{\dot{\omega}} = \mathbf{\tau} - \mathbf{\omega} \times (\mathbf{I} \mathbf{\omega} ) - k_R \mathbf{\omega}
\end{equation}

To perform motion prediction, the equations of motion are discretized for all state variables used in motion prediction as per as per section \ref{sec:formulation}. Parameters are either empirically measured or estimated using nonlinear system identification, as in \cite{mohajerin2019multistep}. Numerical forward integration is then performed using a real-valued variable-coefficient ODE (VODE) solver. The predicted state variables after an interval $\Delta t = 0.01s$ is compared to learning-based methods trained on motion prediction for the same discrete time interval. 

\subsection{Hybrid Models}

On the other hand, we can use all or part of this physics-based model as a component in a hybrid architecture. We develop a series of hybrid models combining fully convolutional Temporal Convolutional Network component(s) with similar design parameters as End2End-TCN and the same total number of parameters. Physics-based components generate forward predictions in a sequential manner by forward integrating some or all of the dynamic system equations outlined in section \ref{physics}. This results in three different Hybrid configurations. Motor-Hybrid uses a TCN component to model the aircraft’s rotor dynamics, generating motor thrust predictions for a given control input. AccelError-Hybrid uses a TCN component to model an additive term to the physics-based state derivative estimates, thus modeling the dynamics that are not captured by the simplified physics-based model. Lastly, Combined-Hybrid uses both TCN components of the models above. 

\begin{figure}
    \centering
    \includegraphics[width=0.45\textwidth]{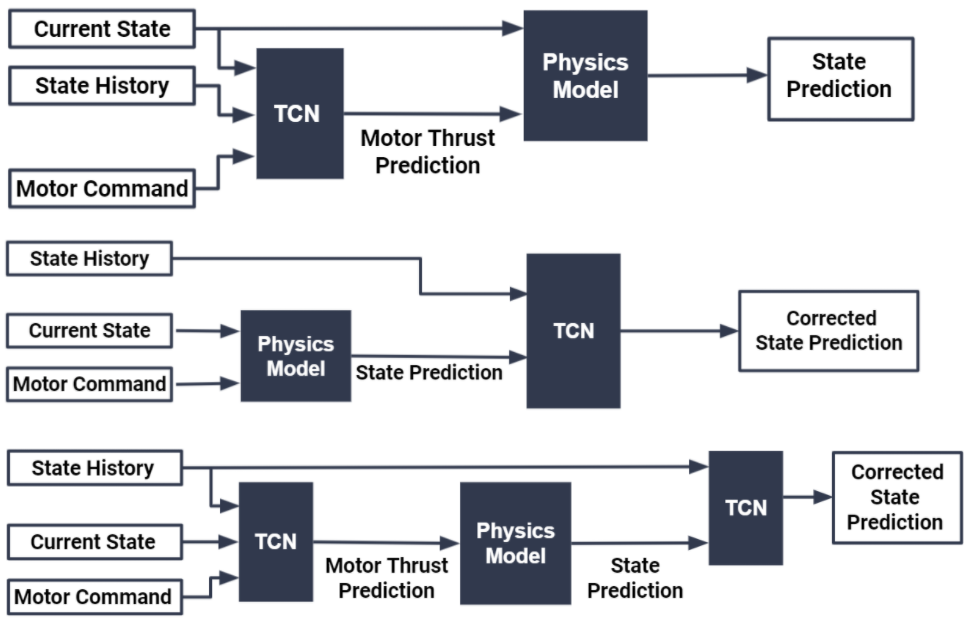}
    \caption{Components of Motor-Hybrid (top), AccelError-Hybrid (middle) and Combined-Hybrid (bottom).}
    \label{fig:hybrids}
\end{figure}

\begin{figure*}[ht]
    \centering
    \includegraphics[width=0.8\textwidth]{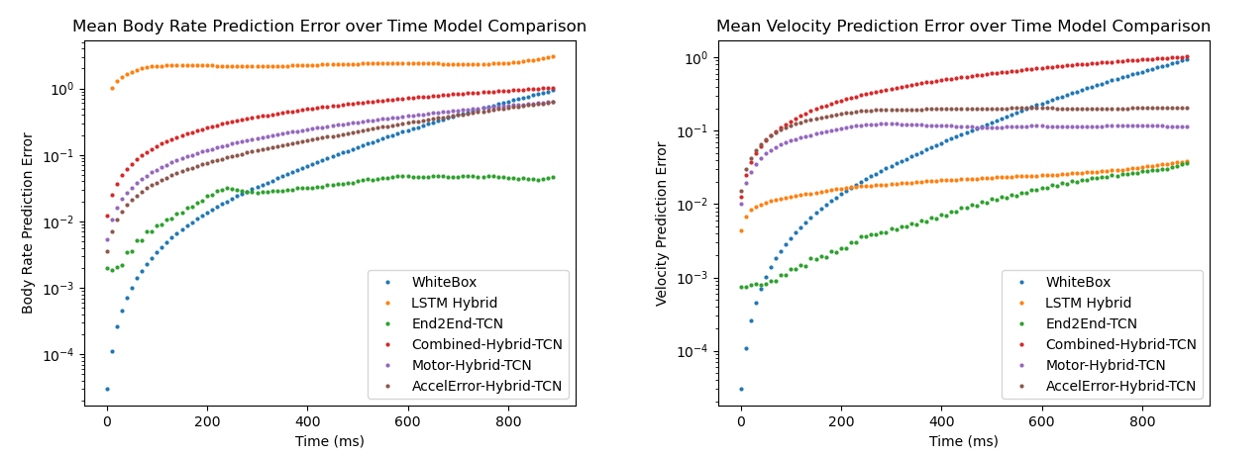}
    \caption{Velocity and body rate prediction errors over time for End2End-TCN (green) and reference models on a log plot.}
    \label{fig:errors}
\end{figure*}
\newpage

\section{Experimental Results}
\label{sec:result}

\subsection{Experimental Design}
We validate this approach and characterize model performance with respect to its prediction accuracy on real quadrotor flights. We evaluate End2End-TCN and several alternative predictive models on the WAVE Laboratory AsTec Pelican Quadrotor Dataset \cite{mohajerin2017modeling}, which utilized sensor fusion across inertial, GNSS, and vision-based systems to collect high-precision quadrotor state estimates. Data are interpolated to report full quadrotor states at a sample rate of 100 Hz. The dataset is comprised of a series of indoor quadrotor flights, bounded within a 5 x 5 x 5 m volume. This mostly includes near-hover flight, pseudo-random rotations and curves in all axes, all within the nominal flight envelope of the AsTec Pelican quadrotor. In total, the dataset consists of 54 flights, with over 1,388,410 total samples of quadrotor telemetry data, 10\% of which is used in the test set for this experiment. 

\subsection{Comparative Study}
To validate the performance of End2End-TCN, we compare its performance in terms of velocity and body rate prediction accuracy with alternative models. This includes the current state-of-the-art result on this dataset, which was achieved by Mohajerin in \cite{mohajerin2019multistep} with an LSTM Recurrent Neural Network Hybrid model to multi-step quadrotor prediction. The model is also compared to a physics-based model, and a series of hybrid models with both TCN and physics-based components, as outlined in section 4.  

\begin{table*}[]
 \centering
 \caption{Summary of multi-step prediction results across 90 time steps (900 ms).}
\adjustbox{max width=\textwidth}{
\begin{tabular}{l|ll|ll|ll}

\toprule
\multicolumn{1}{c}{\multirow{2}{*}{\textbf{Model}}} & \multicolumn{2}{c}{\textbf{MSE Error (t=0.01s)}} & \multicolumn{2}{c}{\textbf{MSE Error (t=0.45s)}}     & \multicolumn{2}{c}{\textbf{MSE Error (t=0.90s)}} \\ 
\cmidrule{2-4} \cmidrule{5-7}
\multicolumn{1}{c}{}                                & \multicolumn{1}{l}{\textbf{Velocity}} & \multicolumn{1}{l}{\textbf{Body Rate}} & \multicolumn{1}{l}{\textbf{Velocity}} & \multicolumn{1}{l}{\textbf{Body Rate}} & \multicolumn{1}{l}{\textbf{Velocity}} & \multicolumn{1}{l}{\textbf{Body Rate}} \\ 
\midrule
Physics-based                                         & \textbf{0.00003}                       & \textbf{0.000572}                       & 0.0892                                 & 0.0981                                  & 0.938                                  & 1.08                                    \\ 
LSTM Hybrid                                           & 0.00441                                & 0.616                                   & 0.0217                                 & 2.30                                    & 0.0384                                 & 3.01                                    \\ 
Motor-Hybrid                                          & 0.0100                                 & 0.00543                                 & 0.115                                  & 0.269                                   & 0.115                                  & 0.632                                   \\ 
AccelError-Hybrid                                     & 0.0153                                 & 0.00356                                 & 0.200                                  & 0.187                                   & 0.205                                  & 0.625                                   \\ 
Combined-Hybrid                                       & 0.0124                                 & 0.0126                                  & 0.178                                  & 0.535                                   & 0.192                                  & 1.02                                    \\ 
\textbf{End2End-TCN}                                  & 0.000735                               & 0.00197                                 & \textbf{0.00881}                       & \textbf{0.0352}                         & \textbf{0.0357}                        & \textbf{0.0464}                         \\ 
\bottomrule
\end{tabular}
}
\end{table*}

We find that End2End-TCN outperforms the current state of the art and all alternative models across nearly the entire 90 step sample (corresponding to 900 ms). The most significant performance improvements are in rotation rates, where the fundamental kinematics rely on current and past quadrotor states. This may indicate that dilated convolutions are better suited for this type of long-term sequence modeling. We find that hybrid models perform significantly worse than the fully convolutional approach. This can mostly be attributed to the difficulty of integrating TCNs with numerically integrated dynamical system equations, which are sequential in nature. Hybrid models that have multiple TCN components, each with a fraction of a single large End2End-TCN, likely suffer due to a fundamental lack of expressive power. 

Lastly, we see that most TCN-based models represent a 2-10x improvement with respect to prediction accuracy when compared to the physics-based model over a longer time horizon, which indicates that these models learn generalizable unmodeled dynamics that have significant temporal effects. We find that TCN model errors typically plateau over time. While a constant acceleration error due to unmodeled disturbances may cause errors growing quadratically over time, End2End-TCN optimizes for accuracy across the flight sample over longer time periods where transient effects may not be statistically relevant.

\subsection{Analyzing Flight Samples}
\begin{figure}[h]
    \centering
    \includegraphics[width=0.48\textwidth]{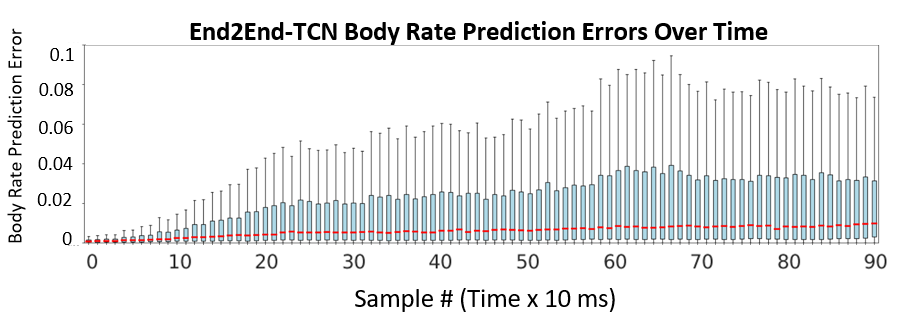}
    \caption{Distribution of End2End-TCN body rate errors over time (Box-whisker plot: median (red line), 2nd and 3rd quartiles (blue box) and range)}
    \label{fig:error_distribution}
\end{figure}

While End2End-TCN makes extremely accurate predictions for a majority of samples, overall accuracy is limited by a long tail in the error distribution as depicted for body rotation rate error in figure~\ref{fig:error_distribution}.

These uncommon but large errors occur at the extremes of the quadrotor's flight envelope. While using an L1-Norm loss function reduces prediction error overall, it constrains the model to learn the simple hover point dynamics, that are more frequent in the training and evaluation datasets. As such, flight samples in more aggressive maneuvers yield predictions that significantly diverges from the ground truth, as in figure 6. We find that samples with errors in the 90th percentile have significantly higher rates of change of position and motor command (i.e. faster and sharper turns). We also find an increase in the variance of pitch and roll angles, indicating that samples taken farther from the hover point of the quadrotor.

\begin{figure*}[ht]
    \centering
    \includegraphics[width=\textwidth]{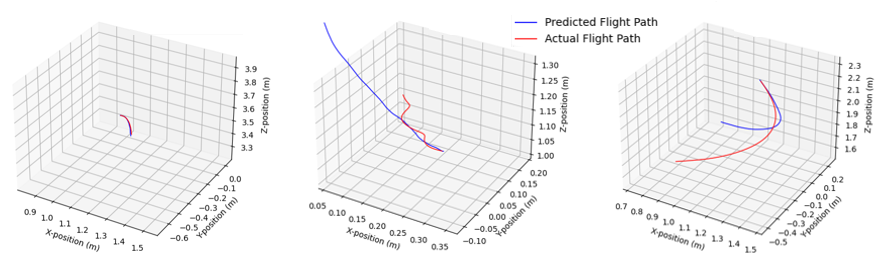}
    \caption{Flight predictions with respect to ground truth for a selection of test samples including low error (frequent) cases and high error (infrequent) cases}
    \label{fig:bad_examples}
\end{figure*}

It is hypothesized that this behavior is largely data-driven. The current dataset, comprised of stable, indoor flight, has few samples in the extreme ranges of the quadrotor's flight envelope. However, in comparison, hybrid models appear to be more robust to these outlier samples. These models have significantly worse mean errors over time but a smaller standard deviation, which indicates that building models with a prior on the system's dynamics may be an effective way to address a lack of data in certain flight modes.

\subsection{Scaling Effects}
One of the main potential benefits of a fully convolutional architecture for quadrotor predictive modeling was its computational efficiency and memory footprint. Thus, we investigate the impact of model size of its predictive modeling performance. Table II shows the validation set accuracy results of End2End-TCN when varying the number of depth layers. Forward pass frequency was calculated on a test set running on a Nvidia GeForce RTX 2080 Ti Graphic Processing Unit (GPU). 

Overall, we find that End2End-TCN retains a significant amount of its predictive ability as the size of the network decreases, particularly for translational velocity. On the other hand, we see significant reductions in body rotation rate prediction accuracy, likely due to the nonlinear nature of these dynamics and their higher sensitivity to disturbances. Similarly, we find that reducing the observation window does not significantly degrade the performance of End2End-TCN. 

One hypothesis for this behavior is that the current model is fundamentally limited by the size of the dataset rather than the size of the model. As demonstrated in language models and other sequence learning tasks \cite{kaplan2020scaling}, performance improvements from increasing model size is fundamentally capped if the size of the dataset does not increase accordingly. There may be additional factors about time-correlated data that make it less susceptible to performance increases from model scale. This view of a data-centric approach for further model scaling is supported by error distributions and the sparsity of data in certain flight modes.

\begin{table}[ht]
\centering
\caption{Prediction error with respect to model size.}
\adjustbox{max width=\textwidth}{
\begin{tabular}{cc|p{0.3in}p{0.3in}|p{0.25in}p{0.25in}}
\toprule
\multirow{2}{*}{\textbf{\begin{tabular}[c]{@{}c@{}}\# of \\ layers\end{tabular}}} & \multirow{2}{*}{\textbf{\begin{tabular}[c]{@{}c@{}}\# of param. \\ / fps (hz) \end{tabular}}} & \multicolumn{2}{c}{\textbf{MSE (t=0.45s)}} & \multicolumn{2}{c}{\textbf{MSE (t=0.90s)}} \\ \cmidrule{3-6} 
 &   & \multicolumn{1}{l}{\textbf{Vel.}} & \multicolumn{1}{l}{\textbf{\begin{tabular}[c]{@{}c@{}}Ang. \\ Vel.\end{tabular}}} & \multicolumn{1}{l}{\textbf{Vel.}} & \multicolumn{1}{l}{\textbf{\begin{tabular}[c]{@{}c@{}}Ang. \\ Vel.\end{tabular}}} \\ 
\midrule
5 & 298,346 / 492.6 & 0.0102 & 0.0387 & 0.0423 & 0.0634 \\ 
8 & 1,166,794 / 383.7 & 0.0088 & \textbf{0.0352} & 0.0357 & \textbf{0.0464} \\ 
10 & 4,640,266 / 302.4 & \textbf{0.0087} & 0.0403 & \textbf{0.0353} & 0.0663 \\ 
12 & 18,517,706 / 243.7 & 0.0148 & 0.0398 & 0.0412 & 0.0654 \\ 
\bottomrule
\end{tabular}
}
\label{tab:E2E_size_table}
\end{table}

\addtolength{\textheight}{-0.75in}   

\subsection{Ablation Studies}
A series of ablation studies is performed on End2End-TCN to validate the model’s detailed design. We first compare a series of alternative architectures. This includes models with varying amounts of regularization layers (Batch Normalization and Dropout) and varying training loss functions (Euclidean, Manhattan, and Weighted Euclidean). The results of the study are summarized in table \ref{table_ablation} for Batch Normalization (BN), Dropout (Drop), Shortened gradient path architecture (SG), Weighted L2-Norm loss function (WL2), and L1-Norm loss (L1).
A crucial element of the design of End2End-TCN is the integration of future control inputs for the multi-step prediction of non-autonomous dynamical systems. In our ablation study, we consider two methods to achieve this. In the baseline model, past quadrotor states, past control inputs, and future control inputs are concatenated into a single model input sequence. We compare this approach to an architecture where only past quadrotor states and control inputs are fed to the first layer, while future control inputs are fed to an intermediate layer for the purposes of shortening their gradient path. 

\begin{figure}[thpb]
  \centering
  \includegraphics[width= 0.48\textwidth]{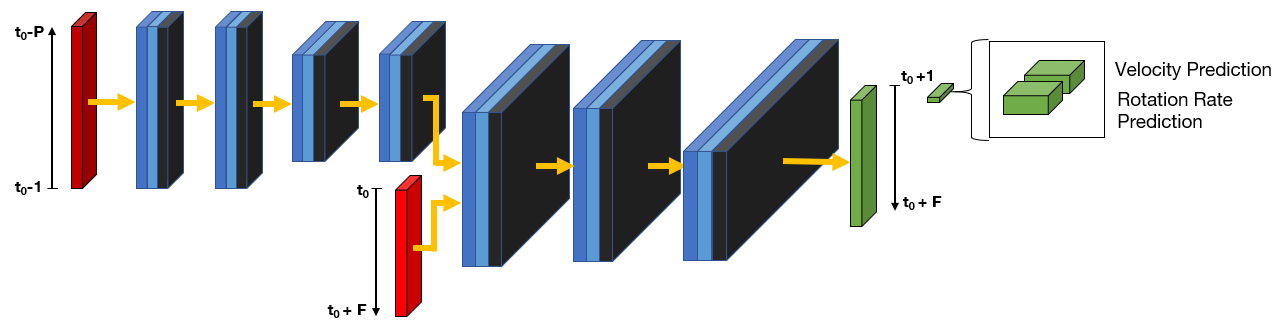}
  
  \caption{Architecture with shortened gradient paths to future control inputs.}
  \label{fig:e2e_compare}
\end{figure}

\begin{table}[ht]
\caption{Architecture Ablation Study}
\label{table_ablation}
\begin{center}
\adjustbox{max width=\textwidth}{
\begin{tabular}{|c|c|c|c|c||c|c|}
\hline
\multirow{2}{*}{\textbf{BN}} & \multirow{2}{*}{\textbf{Drop}} & \multirow{2}{*}{\textbf{SG}} & \multirow{2}{*}{\textbf{WL2}} & \multirow{2}{*}{\textbf{L1}} & \multicolumn{2}{c|}{\textbf{MSE Error}} \\ \cline{6-7} 
                             &                                &                              &                               &                              & \textbf{Velocity}  & \textbf{Body Rate} \\ \hline
            &               &               &               &           & 0.0198        & 0.0715                \\ \hline
\checkmark  &               &               &               &           & 0.0172        & 0.0401                \\ \hline
\checkmark  & \checkmark    &               &               &           & 0.0217        & 0.0433                \\ \hline
\checkmark  &               & \checkmark    &               &           & 0.0329        & 0.0440                \\ \hline
\checkmark  &               &               & \checkmark    &           & 0.0317        & 0.0700               \\ \hline
\checkmark  &               &               &           & \checkmark    & \textbf{0.0158} & \textbf{0.0396}     \\ \hline
\end{tabular}
}
\end{center}
\end{table}
\newpage

Firstly, we see that the alternative architecture performance significantly worse with respect to body rate error when compared to the final model. While this architecture was hypothesized to increase performance by shortening the gradient path to the most important features, namely the last quadrotor state and the control inputs, we see that the number of layers between these features and the output are too few to properly capture the nonlinear rotation dynamics of the quadrotor. Furthermore, reducing or eliminating batch normalization in End2End-TCN decreases performance, as does adding dropout to the model. These results mirror similar conclusions in literature \cite{lea2017temporal}. We also find that the L1-Norm loss function, which is more robust to outlier state errors, leads to better generalization to the test set than do L2 or weighted L2 loss functions. 


\section{Conclusion}
\label{sec:conclusion}
This paper presents a detailed study of the use of Temporal Convolutional Networks for quadrotor state modeling and motion prediction. While classical modeling techniques characterize such robotic systems using prior knowledge of the system's non-autonomous dynamics, we formulate this as a sequence modeling problem by performing discrete multi-step prediction. We segment quadrotor telemetry to train a fully convolutional neural network, End2End-TCN, in a semi-supervised fashion. End2End-TCN outperforms the previous state of the art by 55\% and proves to be more effective than hybrid models and fully physics-based models. We demonstrate that End2End-TCN retains over 95\% of its performance over shorter time intervals when the model is compressed by a factor of 3, and we further characterize model performance with an ablation study and an analysis of predicted flight samples.

This fully convolutional approach to quadrotor modeling is currently limited by the scale and distribution of training data, which is a bottleneck shared by many sequence to sequence models. Collecting data on aggressive quadrotor flight would reduce the model's bias towards hover point dynamics and potentially reduce infrequent low-accuracy prediction samples. Further work is required to ascertain whether this method will generalize to outdoor environments with wind disturbances. Finally, End2End-TCN will be applied in model-based quadrotor control methods to further contextualize its accuracy and computational efficiency. 
\newpage





\bibliographystyle{ieeetr} 
\bibliography{references} 

\end{document}